\documentclass[conference]{IEEEtran}
\IEEEoverridecommandlockouts

\usepackage{cite}
\usepackage{amsmath,amssymb,amsfonts}
\usepackage{algorithmic}
\usepackage{graphicx}
\usepackage{textcomp}
\usepackage{xcolor}
\usepackage{multirow}
\usepackage{float}
\usepackage{hyperref}
\def\BibTeX{{\rm B\kern-.05em{\sc i\kern-.025em b}\kern-.08em
    T\kern-.1667em\lower.7ex\hbox{E}\kern-.125emX}}
\begin{document}

\title{Privacy Risks Analysis and Mitigation in Federated Learning for Medical Images

}

\author{\IEEEauthorblockN{Badhan Chandra Das}
\IEEEauthorblockA{\textit{Knight Foundation School of} \\
\textit{Computing and Information Sciences }\\
Miami, Florida, USA. \\
bdas004@fiu.edu}
\and
\IEEEauthorblockN{M. Hadi Amini}
\IEEEauthorblockA{\textit{Knight Foundation School of} \\
\textit{Computing and Information Sciences}\\
Miami, Florida, USA.\\
moamini@fiu.edu}
\and
\IEEEauthorblockN{Yanzhao Wu}
\IEEEauthorblockA{\textit{Knight Foundation School of} \\
\textit{Computing and Information Sciences}\\
Miami, Florida, USA. \\
yawu@fiu.edu}
\vspace*{-1cm}
}

\maketitle

\begin{abstract}

Federated learning (FL) is gaining increasing popularity in the medical domain for analyzing medical images, which is considered an effective technique to safeguard sensitive patient data and comply with privacy regulations.
However, several recent studies have revealed that the default settings of FL may leak private training data under privacy attacks. Thus, it is still unclear whether and to what extent such privacy risks of FL exist in the medical domain, and if so, ``how to mitigate such risks?''. In this paper, first, we propose a holistic framework for Medical
data Privacy risk analysis and mitigation in Federated
Learning (MedPFL) to analyze privacy risks and develop effective mitigation strategies in FL for protecting private medical data.
Second, we demonstrate the substantial privacy risks of using FL to process medical images, where adversaries can easily perform privacy attacks to reconstruct private medical images accurately.
Third, we show that the defense approach of adding random noises may not always work effectively to protect medical images against privacy attacks in FL, which poses unique and pressing challenges associated with medical data for privacy protection.

\end{abstract}


\section{Introduction}

FL distributes training data across multiple devices, reducing systemic privacy risks by keeping private data decentralized on clients and sharing only gradient updates~\cite{FLmain1}. FL has emerged as a promising learning technique in healthcare by enabling privacy-compliant ML models without sharing raw data among distributed clients like hospitals and clinics~\cite{nguyen2022federated}. 
Medical data, like X-rays and diabetic test reports, contain highly sensitive and personal details, e.g., birth dates that can be used to identify individuals. Breaching such data can lead to severe consequences like stigma, discrimination, or job and insurance loss for patients. Global data protection regulations like HIPPA~\cite{cheng2006health} are in place to safeguard private health information, emphasizing the need for privacy.

\noindent \textbf{Related Works.} Several recent studies~\cite{zhu2019deep,geiping2020inverting,wei2020framework,wei2021gradient,liu2022threats,dahlgaard2022analysing,wei2023securing,wei2023model} have made attempts to recover the private training data and proposed several privacy leakage attacks as Client Privacy Leakage (CPL)~\cite{wei2020framework}, Deep Leakage from Gradients (DLG)~\cite{zhu2019deep}, and Inverting Gradients (GradInv)~\cite{geiping2020inverting}, which demonstrated that the default privacy framework in FL may not be sufficient to prevent privacy leakage in a general FL setting. Introducing random noises in deep learning training can be a viable method to defend against privacy attacks for deep learning models, such as training with Gaussian noise~\cite{abadi2016deep} or Laplacian noise~\cite{melis2015efficient}. However, it lacks in-depth systematic studies of potential risks, privacy attacks, and defense strategies for medical data in FL settings. In the medical domain, Kaissis et. al.~\cite{kaissis2020secure} reported privacy-preservation methods for chest X-rays classification and segmenting CT scans in deep learning training. 
To the best of our knowledge, no frameworks have been proposed to analyze the privacy risks of medical data and its mitigation strategies in the FL environments. Furthermore, we have identified some unique challenges, including complex types, statistical variance, higher dimensionality, and latent pathological information in medical images, which complicate privacy concerns of medical data in FL. 
Therefore, it remains unclear to what extent FL applications in the medical domain are vulnerable to privacy attacks and how to mitigate such privacy risks, especially when FL involves confidential medical data, such as skin cancer images, X-ray images, and MRI scans of patients.

\noindent \textbf{Contributions.} We argue that it is imperative to investigate the privacy risks and develop mitigation strategies to prevent privacy attacks against FL applications in the medical domain.
This paper presents three original contributions. \textit{First,} we present a systematic framework for \textbf{Med}ical data \textbf{P}rivacy risk analysis and mitigation in \textbf{F}ederated \textbf{L}earning (MedPFL). This framework contains a suite of datasets, models, attack methods, defense mechanisms, and evaluation metrics to evaluate the attack and defense with different configurations. \textit{Second,} we demonstrate the severe privacy leakage risks of leveraging FL to analyze medical images. Our empirical evaluations show how an adversary can easily perform privacy attacks and reconstruct patients' private medical data with high reconstruction accuracy. \textit{Third,} we vary the defense configurations to prevent privacy leakage by introducing different levels of random noises in FL with the goal of safeguarding private medical data, which reveals the unique research challenges for protecting private medical data. Experiments are conducted with our MedPFL framework on several benchmark medical image datasets to analyze and mitigate the privacy risks of FL for medical images. We conjecture that this study will draw the attention of the research community toward privacy-preserving techniques in FL for medical applications. The source codes are available on GitHub at \url{https://github.com/mlsysx/MedPFL}.

\section{Framework Overview} \label{section:framework-overview}
\begin{figure}[h]
    \centering
    \includegraphics[scale=.3]{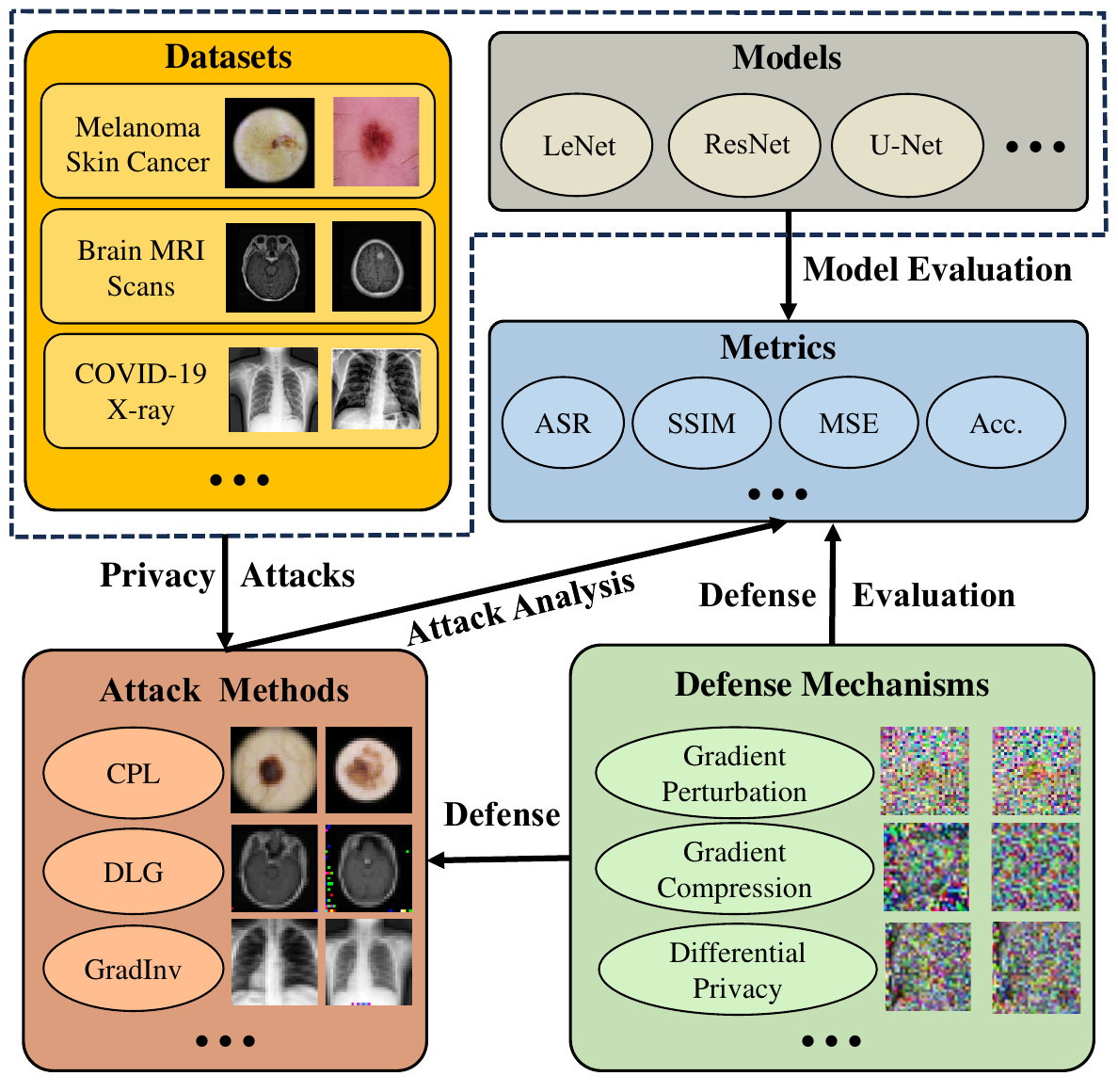}
    \vspace{-2ex}
    \caption{The overview of MedPFL: a framework for \textbf{Med}ical Data \textbf{P}rivacy risk analysis and mitigation in \textbf{F}ederated \textbf{L}earning}
    \label{fig:framework}
    \vspace{-3ex}
\end{figure}

We propose MedPFL, a framework for \textbf{Med}ical data \textbf{P}rivacy risk analysis and mitigation in \textbf{F}ederated \textbf{L}earning. This framework contains five major components to facilitate the evaluation, comparison, and mitigation of privacy risks of processing medical data in FL settings as illustrated in  Figure~\ref{fig:framework}.   
\vspace{-3ex}

\subsection{Framework Components}

\noindent \textbf{Datasets.} Our framework offers publicly available diverse real-world medical datasets, including Melanoma Skin Cancer, COVID-19 X-rays, and Brain Tumor MRI scans, to assess privacy risks. We also provide the APIs to incorporate new datasets in our framework.

\noindent \textbf{Models.}
In our MedPFL framework, we support a variety of popular deep learning models for processing medical data, such as LeNet~\cite{lecun1998gradient} and ResNet~\cite{he2016deep} for medical image classification. These models represent the mainstream DNN architectures for medical data analysis. It can effectively reflect the research challenges by comparing their privacy vulnerabilities and exploring potential mitigation strategies.

\noindent \textbf{Attack Methods.}
For the privacy attacks, we implemented a set of attack methods to analyze the privacy risks in medical data analysis, including CPL~\cite{wei2020framework}, DLG~\cite{zhu2019deep}, and GradInv~\cite{geiping2020inverting}. We leverage these attack methods to perform privacy attacks against different medical datasets and models for analyzing their potential privacy risks. These privacy attack methods are evaluated with widely known evaluation metrics.

\noindent \textbf{Defense Mechanisms.}
In a vanilla FL setting, several defense mechanisms can prevent the different types of privacy attacks, such as gradient perturbation~\cite{wei2020framework}, gradient compression~\cite{wei2020framework}, secure multi-party computation~\cite{goldreich1998secure}, and differential privacy~\cite{li2020multi}. However, for medical data, in-depth studies of the effectiveness and configurations of these defense mechanisms against various privacy attacks are lacking. Our MedPFL framework offers diverse defense mechanisms for exploring their efficacy and the factors impact their performance. We also include a range of evaluation metrics to quantify their privacy protection for medical data.

\noindent \textbf{Evaluation Metrics.} Our framework consists of several core evaluation metrics described as follows. Attack Success Rate (ASR) is the percent of successfully reconstructed samples over the number of attacked samples. Mean Squared Error (MSE) is used to quantify dissimilarity between images, where lower MSE values denote more effective attacks and higher values indicate stronger defense in the context of MedFPL. The Structural Similarity Index Measure (SSIM) assesses the similarity between images, with higher values indicating successful attacks and lower values reflecting stronger defense.

\section{Methodology} \label{section:methodology}

In the healthcare domain, FL involves replicating the ML models from the centralized server and distributing them among a group of clients, which can be clinics, hospitals, and healthcare centers. It typically involves the following steps: 
At step 1, each client receives a global model $\theta$ at round $t$ from the trusted centralized server. We denote it as $\theta(i, t)$ for Client $i$. Each client starts to train its local model $\theta(i, t)$ using its private medical data at step 2. Then in step 3, the local model update (gradients, $\nabla\theta(i, t)$) will be sent to the central server. The central server aggregates local model updates received from clients and updates its global model, which can be performed by using an aggregation strategy, such as Federated Averaging \cite{nilsson2018performance}. This process keeps repeating until the stopping criteria are met, e.g., the number of iterations or desired accuracy.

\vspace{-1ex}

\begin{figure}[h]

    \centering
    \includegraphics[width=1.0\linewidth]{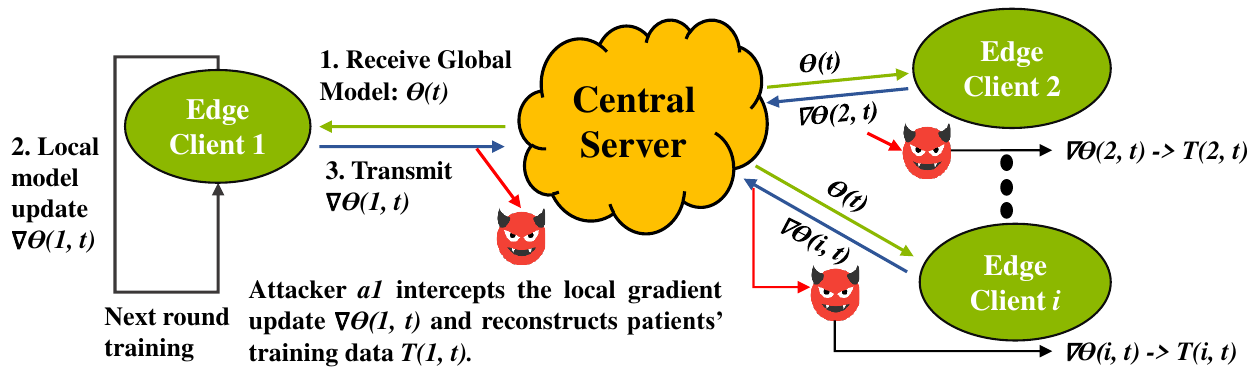}
    \vspace{-4ex}
    \caption{Overview of the Attack Method}
    \label{fig:Proposed_Method}
    \vspace{-2ex}
\end{figure}

\noindent \textbf{Attack Method.}
In Figure \ref{fig:Proposed_Method}, we show how an adversary can intercept the local gradient update and reconstruct the patients' confidential medical data. At step 3, while the local model update (gradients) $\nabla\theta(i, t)$ is shared with a central server, an attacker $a_i$, can intercept the transmission of $\nabla\theta(i, t)$ and acquire the local model update for the corresponding client $i$. Then, this attacker can analyze the periodic local model updates to perform privacy attacks, such as~\cite{wei2020framework,zhu2019deep}, and potentially reconstruct the private data of the client $i$. The privacy attack often starts by initiating dummy data and dummy label that matches the size of the private training data. The attacker then utilizes the dummy data, labels, local model updates (gradients intercepted during step 3), and the shared global model to iteratively update the dummy data and dummy label to reconstruct the client's private training data. This process leverages a reconstruction loss function to minimize the distance between the local gradients on the private data and the dummy gradients on the dummy data and dummy label such that the dummy data gradually becomes close to the private training data to incur privacy leakage.

\noindent \textbf{Defense Mechanism.}
There exist several methods to defend against privacy leakage attacks in the vanilla FL setting, which can be potentially applied to protect the privacy of medical data. Gradient perturbation~\cite{wei2020framework} involves adding a controlled amount of Laplacian or Gaussian noise to the local model update $\nabla\theta(i, t)$ at step 3 in Figure~\ref{fig:Proposed_Method}. 
Noise added to $\nabla\theta(i, t)$ introduces uncertainty in local updates, obscures details, and prevents the accurate reconstruction of private data, e.g., medical images by adversaries. Differential Privacy (DP) is used to add controlled noise to private data, preserving statistical characteristics while hiding individual values~\cite{cynthia2006differential}. This safeguards privacy by preventing the identification of specific data points. Gradient compression~\cite{wei2020framework} and secure multi-party computation~\cite{goldreich1998secure} are also used to prevent privacy leakage in federated learning.

\vspace{-1ex}

\section{Experimental Analysis}~\label{section:experimental-analysis} \vspace{-2ex}

CPL attack has been performed on three representative medical image datasets: Melanoma Skin Cancer, COVID X-ray, and Brain Tumor MRI scans, which contain 10000, 317, and 7022 samples with two, three, and four classes respectively, containing different sizes of images. For example, the Melanoma Skin Cancer dataset consists of images with 300$\times$300 size. COVID X-ray images and brain tumor MRI scans were sized in various shapes. In our experiments, we converted all images into 32$\times$32 size and normalized the data with the mean and standard deviation during data preprocessing. For all the datasets, a 3:1 train-test ratio was maintained. We use 7-layer and 4-layer CNN with a fully connected layer which takes input images with 3 channels following~\cite{wei2020framework} and~\cite{zhu2019deep} for performing the CPL attack and DLG respectively on the medical image datasets. We also perform GradInv~\cite{geiping2020inverting} on Melanoma Skin Cancer images to compare with CPL and DLG.

\noindent \textbf{Attack and Defense Configuration.} 
Both CPL and DLG attack starts with intercepting the local model update $\nabla\theta(i, t)$ for a client $i$ and initializing the dummy image of the same size and the dummy label. Then the dummy image will be iteratively updated to optimize the reconstruction loss to minimize the $L2$ distance between the actual gradient $\nabla\theta(i, t)$ on the private training data and the dummy gradients on the dummy data. L-BFGS optimization method is employed as suggested by~\cite{zhu2019deep} and \cite{wei2020framework} for performing the privacy attack.
The GradInv attack basically works on the gradients, $\nabla\theta(i, t)$ of the local training data to reconstruct the original images with a network formed with fully connected layers~\cite{geiping2020inverting}.
Given that the non-zero gradient condition is met, it iteratively analyzes the $\nabla\theta(i, t)$ to optimize the loss function (cosine similarity) to reconstruct the network’s input. 
As suggested by \cite{geiping2020inverting}, we employed Adam as an optimization algorithm.

In this research, gradient perturbation is performed as a defense mechanism to prevent privacy leakage attacks. Here, we insert a controlled amount of Laplacian noise with zero means to the gradients while transmitting $\nabla\theta(i, t)$ to the central server. Noise added to $\nabla\theta(i, t)$ introduces uncertainty in local updates, hides details, and prevents accurate reconstruction of private data of medical images by adversaries. Our empirical analysis revealed that the default amount of noise may not always provide sufficient defense for medical images. Thus, we vary noise levels to identify stronger defense settings.

\begin{table}[h]

\caption{Performance Comparison of CPL attack on all three medical datasets for 100 images.}
\centering
\scalebox{0.8}{
\begin{tabular}{|c|c|c|c|c|}
\hline

\textbf{Dataset}                                                                 & \textbf{ASR} & \textbf{\begin{tabular}[c]{@{}c@{}}Avg. \\ SSIM\end{tabular}} & \textbf{\begin{tabular}[c]{@{}c@{}}Avg. \\ MSE\end{tabular}} & \textbf{\begin{tabular}[c]{@{}c@{}}Avg, Duration \\ per Image (in seconds)\end{tabular}} \\ \hline
\textbf{\begin{tabular}[c]{@{}c@{}}Melanoma Skin Cancer \end{tabular}} & 49$\%$                                                                   & 0.50                                                                         & 0.0762                                                                      & 50.847                                                                                   \\ \hline
\textbf{\begin{tabular}[c]{@{}c@{}}COVID-19 X-ray Images\end{tabular}}              & 55$\%$                                                                     & 0.55                                                                         & 0.0641                                                                      & 92.245                                                                                  \\ \hline
\textbf{\begin{tabular}[c]{@{}c@{}}Brain Tumor MRI Scans\end{tabular}}             & 76$\%$                                                                     & 0.75                                                                         & 0.0586                                                                      & 103.12                                                                                   \\ \hline

\end{tabular}
}
\vspace{-2.5ex}
\label{tab:exp_results1}
\end{table}

\begin{figure}[h]
    \centering
   
    \includegraphics[scale=0.5]{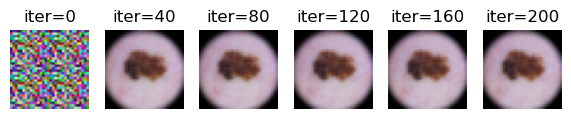}
    \vspace{-1.8ex}
    
    \caption{CPL attack on Melanoma Skin Cancer Dataset}
    \label{fig:Attack_result1}
    
    \includegraphics[scale=0.5]{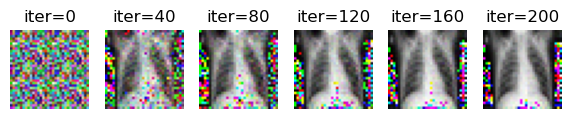}
    \vspace{-1.8ex}
    \caption{CPL attack on COVID X-ray Images Dataset}
    
    \label{fig:Attack_result2}
    
\vspace{-4ex}
\end{figure}

\noindent \textbf{Performance of Attack Methods.} 
We show the performance comparison of the CPL attack framework on different medical image datasets in Table~\ref{tab:exp_results1} for randomly sampled 100 images. We observe that the SSIM measures (in the 3rd column) are 0.5, 0.55, and 0.75 for Skin Cancer dataset, COVID-19 dataset, and Brain MRI dataset respectively. When computing ASR, an SSIM value of 0.9 or higher between the original private images and the attack-reconstructed image has been considered as a successful attack for all methods in this study. From the ASR values (2nd column) we observe that around 50$\%$ 55$\%$ and 75$\%$ images have been reconstructed successfully for the Skin Cancer dataset, COVID-19 dataset, and Brain MRI dataset respectively. We also compare their execution time to perform the privacy attacks in all cases (5th column in Table \ref{tab:exp_results1}). Figure~\ref{fig:Attack_result1} and ~\ref{fig:Attack_result2}  
visually show the intermediate reconstructed images by a successful CPL attack on Skin Cancer and COVID-19 datasets for 200 iterations respectively, where the private medical images can be accurately reconstructed with a very small noise.

\vspace{-2.5ex}

\begin{table}[!h]
\tiny
\centering
\caption{Performance Comparison of three attack methods on Melanoma Skin Cancer dataset for 100 images.}
\vspace{-2ex}
\centering
\begin{tabular}{|c|c|c|c|c|c|}
\hline

\textbf{Method}                                                                & \textbf{Model}                                                             & \textbf{ASR} & \textbf{\begin{tabular}[c]{@{}c@{}}Avg. \\ SSIM\end{tabular}} & \textbf{\begin{tabular}[c]{@{}c@{}}Avg. \\ MSE\end{tabular}} & \textbf{\begin{tabular}[c]{@{}c@{}}Avg, Duration \\ per Image (in seconds)\end{tabular}} \\ \hline
\multirow{2}{*}{\begin{tabular}[c]{@{}c@{}}GradInv\end{tabular}} & \begin{tabular}[c]{@{}c@{}}ResNet-18  Untrained\end{tabular}              & 62\%                                                                   & 0.9029                                                                       & 0.9434                                                                      & 6019.84                                                                                  \\ \cline{2-6} 
                                                                               & \begin{tabular}[c]{@{}c@{}}ResNet-18 Trained \\ with ImageNet\end{tabular} & \textbf{82\%}                                                                   & \textbf{0.9271}                                                                       & \textbf{0.8027}                                                                      & \textbf{4880.58}                                                                                  \\ \hline
CPL                                                                            & CNN                                                                        & 49$\%$                                                                     & 0.50                                                                         & 0.0762                                                                      & 50.847                                                                                    \\ \hline
DLG                                                                            & CNN                                                                        & 47$\%$                                                                     & 0.48                                                                         & 0.0957                                                                      & 60.128                                                                                   \\ \hline
\end{tabular}

\label{tab:exp_results2}
\vspace{-1.5ex}
\end{table}

\begin{figure}[h]
    \centering
    
    \includegraphics[scale=.33]{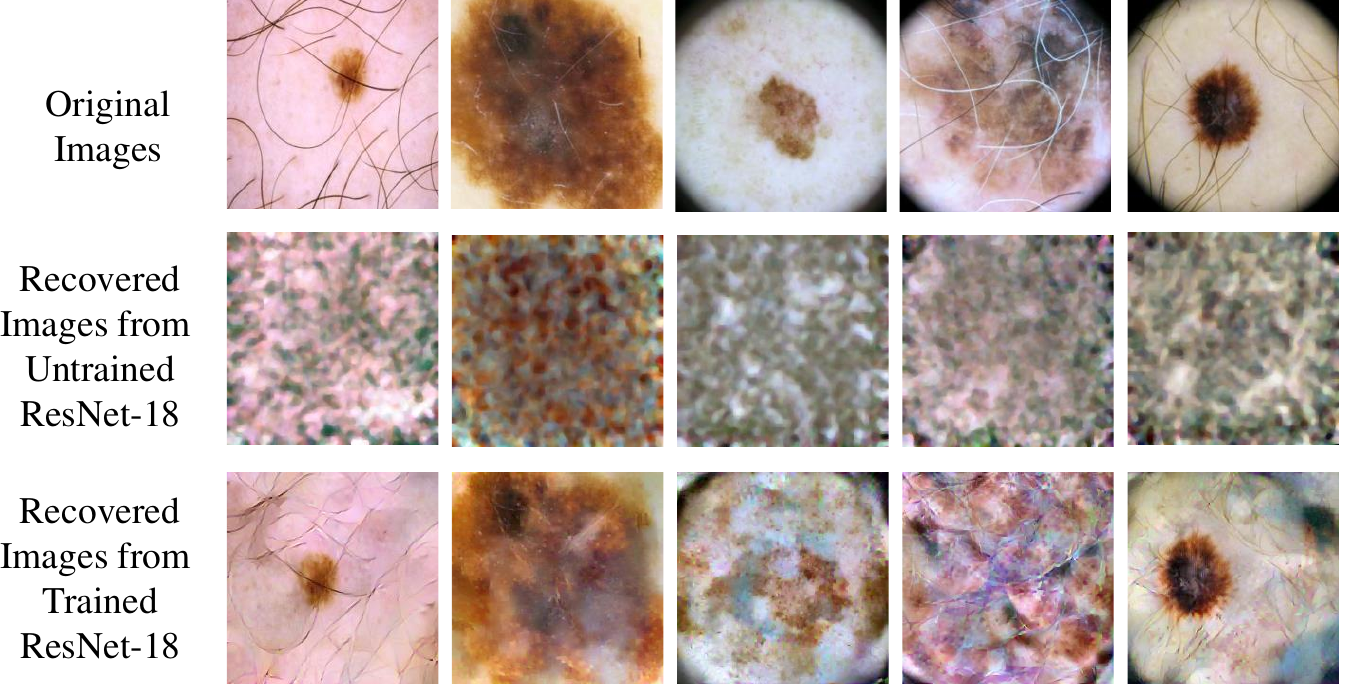}
    \vspace{-2ex}
    \caption{GradInv attack recovered images by untrained and trained ResNet-18 
    }
    \label{fig:inv_grad_attack}
    \vspace{-1ex}
\end{figure}

We also compare GradInv on the Skin Cancer dataset on randomly sampled 100 images for both ResNet-18 \cite{he2016deep} trained and untrained versions with CPL and DLG attacks. We show the performance of GradInv in the first two rows of Table~\ref{tab:exp_results2}. We observed that the performance of GradInv is better on the trained ResNet-18 than the untrained one in terms of all evaluation metrics (higher ASR, higher SSIM, and lower MSE). The GradInv attack on the trained ResNet-18 also takes less time to reconstruct the original private medical images than the untrained ResNet-18. Figure~\ref{fig:inv_grad_attack} visually demonstrates the attack performance by the GradInv on Melanoma Skin Cancer Dataset for 24,000 iterations which supports our observations from Table~\ref{tab:exp_results2}. Comparing the performance of CPL and DLG (bottom two rows in Table \ref{tab:exp_results2}), we observe that GradInv can reconstruct high-quality images by analyzing the models which are well-trained, such as trained ResNet-18. Though GradInv takes much longer time than CPL and DLG, the performance is much better in terms of reconstruction quality as evaluated by ASR, SSIM and MSE. 
\vspace{-2ex}

\vspace{-1.5ex}

\begin{table}[!h]

\vspace{-2ex}
\caption{CPL attack and defense performance for different noise levels on three benchmark datasets.}
\vspace{-1.5ex}
\centering
\scalebox{.75}{
\begin{tabular}{|cc|cc|ccc|}
\hline
\multicolumn{2}{|c|}{\multirow{2}{*}{\textbf{Dataset}}} & \multicolumn{2}{c|}{\textbf{CPL Attack}}              & \multicolumn{3}{c|}{\textbf{Defense}}                                                                                                                                                                                                                              \\ \cline{3-7} 
\multicolumn{2}{|c|}{}                                  & \multicolumn{1}{c|}{\textbf{MSE}} & \textbf{SSIM} & \multicolumn{1}{c|}{\textbf{\begin{tabular}[c]{@{}c@{}}Noise \\ Levels\end{tabular}}} & \multicolumn{1}{c|}{\textbf{MSE}}                                                              & \textbf{SSIM}                                                             \\ \hline
\multicolumn{1}{|c|}{1.} & Melanoma Skin Cancer & \multicolumn{1}{c|}{0.0762}      & 0.50        & \multicolumn{1}{c|}{\begin{tabular}[c]{@{}c@{}}100\\ 200\\ 300\\ \textbf{400}\end{tabular}}    & \multicolumn{1}{c|}{\begin{tabular}[c]{@{}c@{}}0.1306\\ 0.1468\\ 0.1497\\ \textbf{0.1503}\end{tabular}} & \begin{tabular}[c]{@{}c@{}}0.0154\\ 0.0131\\ 0.0121\\ \textbf{0.0101}\end{tabular} \\ \hline
\multicolumn{1}{|c|}{2.} & COVID X-ray Images           & \multicolumn{1}{c|}{0.0641}      & 0.55        & \multicolumn{1}{c|}{\begin{tabular}[c]{@{}c@{}}100\\ 200\\ 300\\ \textbf{400}\end{tabular}}    & \multicolumn{1}{c|}{\begin{tabular}[c]{@{}c@{}}0.0013\\ 0.0157\\ 0.0206\\ \textbf{0.0578}\end{tabular}} & \begin{tabular}[c]{@{}c@{}}0.9815\\ 0.7410\\ 0.7000\\ \textbf{0.4605}\end{tabular} \\ \hline
\multicolumn{1}{|c|}{3.} & Brain Tumor MRI Scans        & \multicolumn{1}{c|}{0.0586}      & 0.75        & \multicolumn{1}{c|}{\begin{tabular}[c]{@{}c@{}}100\\ 200\\ 300\\ \textbf{400}\end{tabular}}    & \multicolumn{1}{c|}{\begin{tabular}[c]{@{}c@{}}0.0207\\ 0.0686\\ 0.0300\\ \textbf{0.1705}\end{tabular}} & \begin{tabular}[c]{@{}c@{}}0.6657\\ 0.2383\\ 0.3661\\ \textbf{0.0699}\end{tabular} \\ \hline

\end{tabular}
}

\label{tab:exp_results3}
\vspace{-2ex}
\end{table}

\begin{figure}[h]
    \vspace{-2ex}
    \centering
    
    \includegraphics[scale=0.33]{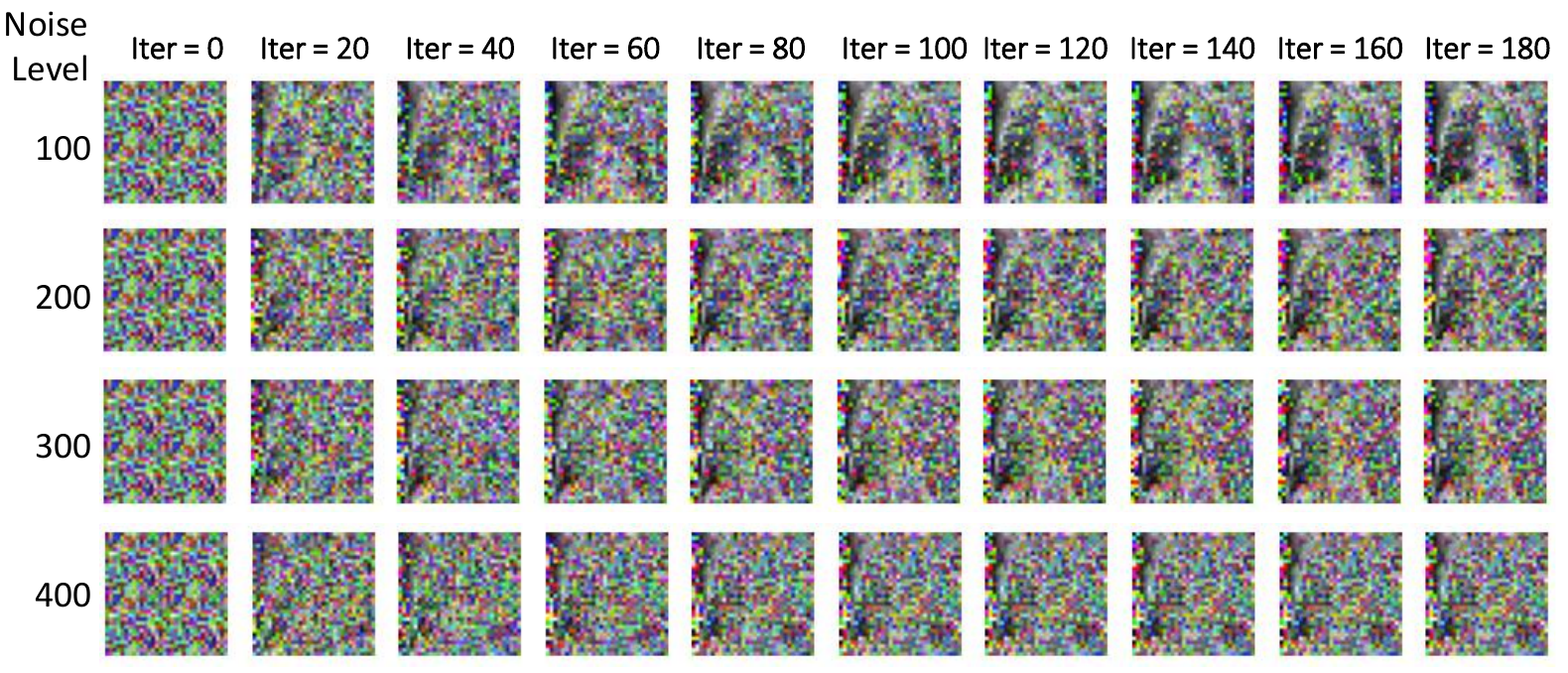}
    \vspace{-5ex}
    \caption{Defense against CPL attack on COVID X-ray Dataset}
    \label{fig:defense_result2}
    
    \vspace{-2ex}

\end{figure}

\noindent \textbf{Performance of Defense Mechanisms.}
The goal of defending privacy attacks is to maximize the dissimilarity (e.g., MSE) and minimize the similarity (e.g., SSIM) between the original training images and the attack reconstructed images. Table~\ref{tab:exp_results3}, shows as we increase the level of Laplacian noise the defense becomes stronger. The addition of random noise to $\nabla\theta(i, t)$ increases the difficulty of extracting private information from the local gradients. Figure~\ref{fig:defense_result2} visualizes the outcome of the defense mechanism for different noise levels for the COVID X-ray dataset. When the noise level is relatively low, such as 100, private information can still be extracted under the CPL attack (first row of Figure~\ref{fig:defense_result2}), which indicates that adding random noise may not always provide adequate privacy protection for medical data in FL.
We found similar observations consistently for the other two medical datasets.

\section{Conclusion} \label{section:conclusion}

This paper introduces MedPFL, a framework for privacy risk analysis and mitigation in FL with medical images. We highlight the extensive privacy risks in processing medical data in FL, emphasizing the need for stronger defenses against adversarial attacks. We use different levels of random noise for defense, noting that while higher noise levels enhance privacy protection, they may not suffice. We demonstrate real-world scenarios of privacy attacks on medical images across benchmark datasets, which further illustrate the critical challenges of mitigating privacy risks in FL in the medical domain. Our future work will explore additional privacy attacks and innovative techniques to protect medical data in FL.

\bibliographystyle{ieeetr}

\bibliography{ref}

\begin{thebibliography}{10}

\bibitem{FLmain1}
B.~McMahan {\em et~al.}, ``Communication-efficient learning of deep networks from decentralized data,'' in {\em Artificial intelligence and statistics}, pp.~1273--1282, PMLR, 2017.

\bibitem{nguyen2022federated}
D.~C. Nguyen {\em et~al.}, ``Federated learning for smart healthcare: A survey,'' {\em ACM Computing Surveys (CSUR)}, vol.~55, no.~3, pp.~1--37, 2022.

\bibitem{cheng2006health}
V.~S. Cheng {\em et~al.}, ``Health insurance portability and accountability act {(HIPPA)} compliant access control model for web services,'' {\em International Journal of Healthcare Information Systems and Informatics (IJHISI)}, vol.~1, no.~1, pp.~22--39, 2006.

\bibitem{zhu2019deep}
L.~Zhu {\em et~al.}, ``Deep leakage from gradients,'' {\em Advances in neural information processing systems}, vol.~32, 2019.

\bibitem{geiping2020inverting}
J.~Geiping {\em et~al.}, ``Inverting gradients-how easy is it to break privacy in federated learning?,'' {\em Advances in Neural Information Processing Systems}, vol.~33, pp.~16937--16947, 2020.

\bibitem{wei2020framework}
W.~Wei {\em et~al.}, ``A framework for evaluating client privacy leakages in federated learning,'' in {\em Computer Security-25th European Symposium on Research in Computer Security, ESORICS 2020, Guildford, UK, September 14--18, 2020, Proceedings, Part I 25}, pp.~545--566, Springer, 2020.

\bibitem{wei2021gradient}
W.~Wei {\em et~al.}, ``Gradient-leakage resilient federated learning,'' in {\em 2021 IEEE 41st International Conference on Distributed Computing Systems (ICDCS)}, pp.~797--807, IEEE, 2021.

\bibitem{liu2022threats}
P.~Liu, X.~Xu, and W.~Wang, ``Threats, attacks and defenses to federated learning: issues, taxonomy and perspectives,'' {\em Cybersecurity}, vol.~5, no.~1, pp.~1--19, 2022.

\bibitem{dahlgaard2022analysing}
M.~E. Dahlgaard {\em et~al.}, ``Analysing the influence of attack configurations on the reconstruction of medical images in federated learning,'' {\em arXiv preprint arXiv:2204.13808}, 2022.

\bibitem{wei2023securing}
W.~Wei {\em et~al.}, ``Securing distributed {SGD} against gradient leakage threats,'' {\em IEEE Transactions on Parallel and Distributed Systems}, vol.~34, no.~7, pp.~2040--2054, 2023.

\bibitem{wei2023model}
W.~Wei, K.-H. Chow, F.~Ilhan, Y.~Wu, and L.~Liu, ``Model cloaking against gradient leakage,'' in {\em 2023 IEEE International Conference on Data Mining (ICDM)}, 2023.

\bibitem{abadi2016deep}
M.~Abadi {\em et~al.}, ``Deep learning with differential privacy,'' in {\em Proceedings of the 2016 ACM SIGSAC conference on computer and communications security}, pp.~308--318, 2016.

\bibitem{melis2015efficient}
L.~Melis {\em et~al.}, ``Efficient private statistics with succinct sketches,'' {\em arXiv preprint arXiv:1508.06110}, 2015.

\bibitem{kaissis2020secure}
G.~A. Kaissis {\em et~al.}, ``Secure, privacy-preserving and federated machine learning in medical imaging,'' {\em Nature Machine Intelligence}, vol.~2, no.~6, pp.~305--311, 2020.

\bibitem{lecun1998gradient}
Y.~LeCun {\em et~al.}, ``Gradient-based learning applied to document recognition,'' {\em Proceedings of the IEEE}, vol.~86, no.~11, pp.~2278--2324, 1998.

\bibitem{he2016deep}
K.~He {\em et~al.}, ``Deep residual learning for image recognition,'' in {\em Proceedings of the IEEE conference on computer vision and pattern recognition}, pp.~770--778, 2016.

\bibitem{goldreich1998secure}
O.~Goldreich, ``Secure multi-party computation,'' {\em Manuscript. Preliminary version}, vol.~78, no.~110, 1998.

\bibitem{li2020multi}
X.~Li {\em et~al.}, ``Multi-site fmri analysis using privacy-preserving federated learning and domain adaptation: Abide results,'' {\em Medical Image Analysis}, vol.~65, p.~101765, 2020.

\bibitem{nilsson2018performance}
A.~Nilsson {\em et~al.}, ``A performance evaluation of federated learning algorithms,'' in {\em Proceedings of the second workshop on distributed infrastructures for deep learning}, pp.~1--8, 2018.

\bibitem{cynthia2006differential}
D.~Cynthia, ``Differential privacy in automata, languages and programming, bugliesi michele, preneel bart, sassone vladimiro, and wegener ingo,'' 2006.

\end{thebibliography}
\end{document}